\documentclass[final,5p,times,twocolumn]{elsarticle} 

\usepackage{amssymb}
\usepackage{amsmath}
\usepackage[dvipsnames]{xcolor}
\usepackage{booktabs}
\usepackage{comment}

\usepackage[breaklinks]{hyperref}

\journal{Pattern Recognition}

\begin{document}

\begin{frontmatter}



\title{Poseidon: A ViT-based Architecture for Multi-Frame Pose Estimation with Adaptive Frame Weighting and Multi-Scale Feature Fusion}




\author[label1]{Cesare Davide Pace\corref{cor1}}
\ead{cesaredavide.pace@unicas.it}
\cortext[cor1]{Corresponding author}

\author[label2,label3]{Alessandro Marco De Nunzio}

\affiliation[label1]{organization={Department of Electrical and Information Engineering, University of Cassino and Southern Lazio},
            country={Italy}}

\affiliation[label2]{organization={Department of Research and Development, LUNEX},
            country={Luxembourg}}

\affiliation[label3]{organization={Luxembourg Health \& Sport Sciences Research Institute ASBL},
            country={Luxembourg}}

\author[label1]{Claudio De Stefano}
\author[label1]{Francesco Fontanella}

\author[label1]{Mario Molinara}

\begin{abstract}
Human pose estimation, a vital task in computer vision, involves detecting and localising human joints in images and videos. While single-frame pose estimation has seen significant progress, it often fails to capture the temporal dynamics for understanding complex, continuous movements. We propose Poseidon, a novel multi-frame pose estimation architecture that extends the ViTPose model by integrating temporal information for enhanced accuracy and robustness to address these limitations. Poseidon introduces key innovations: (1) an Adaptive Frame Weighting (AFW) mechanism that dynamically prioritises frames based on their relevance, ensuring that the model focuses on the most informative data; (2) a Multi-Scale Feature Fusion (MSFF) module that aggregates features from different backbone layers to capture both fine-grained details and high-level semantics; and (3) a Cross-Attention module for effective information exchange between central and contextual frames, enhancing the model’s temporal coherence. The proposed architecture improves performance in complex video scenarios and offers scalability and computational efficiency suitable for real-world applications.
Our approach achieves state-of-the-art performance on the PoseTrack21 and PoseTrack18 datasets, achieving mAP scores of 88.3 and 87.8, respectively, outperforming existing methods. The
code and models are available at \url{https://github.com/CesareDavidePace/poseidon}
\end{abstract}



\begin{keyword}
Human Pose Estimation \sep Multi-Frame Approach \sep Temporal Context \sep ViTPose \sep PoseTrack21 \sep PoseTrack18
\end{keyword}

\end{frontmatter}




\section{Introduction}
\label{sec:intro}
Human pose estimation, the task of detecting and localizing human joints in images or videos, is a cornerstone of computer vision with wide-ranging applications such as human-computer interaction~\cite{salti2008real}, action recognition~\cite{song2021human}, sports analytics~\cite{baumgartner2023monocular}, healthcare monitoring~\cite{difini2021human}, and augmented reality~\cite{marchand2015pose}. This field is essential for understanding human movement and behaviour, enabling machines to interpret human actions effectively.

Traditional single-frame pose estimation techniques isolate each frame, neglecting the crucial temporal dynamics for accurately capturing complex and continuous human movements in videos. On the other hand, multi-frame pose estimation incorporates video sequence analysis, providing advantages such as improved motion continuity and better handling of occlusions, thus outperforming single-frame methods~\cite{liu2021deep, bertasius2019learning}. Integrating information across multiple frames enhances human pose understanding over time, yielding greater robustness and accuracy, especially during intricate activities\cite{liu2021deep, liu2022temporalfeaturealignmentmutual_famipose, feng2023mutual}.

Despite significant advancements in multi-frame pose estimation, challenges remain, such as ensuring temporal consistency, achieving computational efficiency, and effectively handling occlusions in multi-person scenarios. A variety of architectural approaches have been explored to address these challenges. Heatmap-based~\cite{liu2021deep, liu2022temporal, feng2023mutual} and regression-based~\cite{he2024video, mao2022poseur} methods leverage feature extraction from frame sequences for enhanced detection, while approaches involving diffusion models~\cite{feng2023diffpose} and optical flow~\cite{Pfister_2015_ICCV} capture motion dynamics. Additionally, architectures such as LSTMs~\cite{8756597, Artacho_2020_CVPR} and 3D CNNs~\cite{Wang_2020_CVPR} have been employed to bolster temporal coherence in sequences.

While multi-frame methods provide substantial improvements, single-frame pose estimation has advanced notably, especially with Vision Transformer (ViT) architectures~\cite{dosovitskiy2021imageworth16x16words}. Among these, ViTPose~\cite{xu2022vitpose} stands out as a state-of-the-art single-frame pose estimator, demonstrating excellent spatial awareness and generalization. 
In this paper, to capitalize on the spatial strengths of ViTPose while incorporating temporal information, we propose adapting this architecture for multi-frame pose estimation. We aim to merge ViTPose's superior spatial feature extraction capabilities with the temporal coherence of multi-frame techniques to address limitations related to dynamic movements and occlusions. To achieve this, we introduce Poseidon, an architecture that extends the ViTPose model for multi-frame pose estimation. Poseidon integrates temporal context through key innovations that preserve the spatial strengths of the original model.

The main contributions of our work include:
\begin{itemize}
\item A AWF mechanism, which dynamically prioritizes frames based on relevance. This enhances the model's ability to focus on the most informative frames, boosting pose estimation accuracy in complex sequences.
\item A MSFF module that combines features from different layers of the backbone network, enriching the model's representation by capturing fine-grained details and high-level semantic information.
\item A cross-attention module facilitates effective information exchange between center and context frames, thereby improving the model's utilization of temporal context.
\item Overcoming state-of-the-art performance in challenging real-world data sets such as PoseTrack21~\cite{posetrack21} and PoseTrack18~\cite{posetrack18}, underscoring the effectiveness of our approach in multiframe pose estimation.
\item Evaluate the generalization capability of our Poseidon model; we further tested the model trained on the PoseTrack21 dataset directly on the Sub-JHMDB dataset.

\end{itemize}

\section{Related Work}\label{sec:relatedwork}
The transition from single-frame to multi-frame pose estimation has o\-pened new avenues for improving accuracy and temporal consistency in human motion analysis. This section provides a comprehensive overview of the latest key architectural approaches in multi-frame pose estimation that we have been dealing with.

\subsection{Regression-Based Methods}
Regression-based methods for pose estimation operate by directly predicting the coordinates of keypoints from input frames. These methods offer simplicity in design but often need help with the spatial ambiguity of human poses in video sequences. In~\cite{he2024video}, the authors propose Decoupled Space-Time Aggregation (DSTA) architecture, a unique approach to human pose regression in video sequences by separating human joints' spatial and temporal dynamics. Instead of regressing joint coordinates from the output feature maps of a CNN backbone as in traditional regression models, DSTA transforms the backbone's output into a sequence of joint-specific tokens. Each token represents the feature embedding of a single joint, allowing the model to capture spatiotemporal dependencies flexibly. By decoupling space and time, DSTA avoids the issue of conflating these two types of information, which is a common problem in earlier regression methods. This separation enables the architecture to achieve high performance and despite the improvement, heatmap-based methods achieve better performance than regression-based ones.

\subsection{Diffusion-based Methods}
Diffusion models have recently gained traction in computer vision tasks, and their application to human pose estimation introduces a new perspective for modelling spatial and temporal dynamics in video sequences. One of the pioneering approaches in this area is the SpatioTemporal Diffusion Model for Pose Estimation (DiffPose)~\cite{feng2023diffpose}, which extends the diffusion model framework to tackle the challenge of multi-based human pose estimation by treating it as a generative task.

DiffPose reformulates pose estimation as a conditional generative process of keypoint-wise heatmaps. The model consists of two key stages: a forward diffusion stage, where Gaussian noise is progressively added to the ground truth heatmaps, and a reverse denoising stage, where a Pose-Decoder progressively reconstructs the original heatmap from the noisy input. This approach allows DiffPose to leverage both spatial and temporal information innovatively.

\subsection{Heatmap-Based Methods}
Heatmap-based methods have emerged as a popular approach for pose estimation, offering significant improvements in accuracy by predicting per-pixel likelihoods of keypoints in the form of heatmaps. Compared to regression-based methods, they usually achieve higher performance; these methods are used when precision is the priority in keypoint detection. 
In~\cite{liu2021deep}, the authors propose a DCPose architecture incorporating bidirectional temporal information from consecutive frames to enhance pose estimation in video sequences. DCPose~\cite{liu2021deep} introduces a multi-step refinement process. It encodes spatial and temporal contexts into localized search areas, computes pose residuals, and refines heatmap predictions using modules like Pose Temporal Merger, Pose Residual Fusion, and Pose Correction Network.
A  contribution has been the Temporal Difference Learning based on Mutual Information (TDMI) in~\cite{feng2023mutual}. TDMI introduces a novel way to model motion contexts and enhance pose estimation through two key innovations: temporal difference encoding, which models motion contexts by computing feature differences between video frames across multiple stages and representation disentanglement, which is designed to distil task-relevant motion features by separating useful features from noisy constituents in the motion representation. 
Another approach, presented in~\cite{wu2024joint}, proposes Joint-Motion Mutual Learning for Pose Estimation (JM-Pose). JM-Pose introduces two core innovations for pose estimation in video sequences. First, the context-aware joint learner module captures joint-specific features using modulated deformable operations guided by initial heatmaps, enabling precise keypoint localization even in challenging conditions like occlusion and motion blur. This module ensures joint positional accuracy by directly incorporating heatmap information while adapting to changing visual contexts. Second, the progressive joint motion mutual learning mechanism dynamically exchanges and refines information between local joint features and global motion flow, allowing the model to capture both joint-level and motion dynamics simultaneously. An information orthogonality objective is introduced to fully explore the diversity of joint features and motion information.

Unlike previous methods that rely solely on temporal consistency or spatial refinement independently, Poseidon's architecture integrates both aspects by combining the spatial strengths of ViTPose with temporal coherence modules. Specifically, the AFW mechanism distinguishes our approach by dynamically prioritizing frames based on their relevance. Additionally, the MSFF module combines fine-grained and high-level semantic features across scales, ensuring robust keypoint localization at multiple levels of detail. Furthermore, our cross-attention module establishes effective information exchange between center and context frames, directly addressing challenges of motion blur, occlusions, and temporal consistency.

\section{Methods}\label{sec:methods}

This section describes the Poseidon architecture for multi-frame human pose estimation. Poseidon builds upon the ViTPose model~\cite{xu2022vitpose}, adapting it for temporal analysis by incorporating an AFW, an MSFF, and a cross-attention mechanism. This architecture is designed to leverage spatial and temporal information, enabling precise and robust pose estimation in video sequences. 

Poseidon follows a top-down paradigm, where human keypoints are estimated by first detecting the bounding box (bbox) of the subject. The bbox serves as a region of interest in which pose estimation is conducted. 

Given that Poseidon is designed for multi-frame pose estimation, it processes sequences of frames from a video. For each sequence, given the bbox of the person in the central frame, the bbox is then enlarged to account for motion across frames and applied to the other frames in the sequence. This approach should ensure that the person remains within the region of interest throughout the temporal sequence, capturing all possible movements within the bbox.

Formally, let $\mathcal{F} = \{f_{t+k}\}_{k=-T}^{T}$ represent a sequence of $2T + 1$ frames centered around frame $ f_t $. Given the subject's bbox $\mathbf{b}_t$ at the central frame $f_t$, the enlarged bbox $\hat{\mathbf{b}}_t$ is defined as:
\[
\hat{\mathbf{b}}_t = \delta\mathbf{b}_t
\]
where $\delta$ is a margin that accounts for potential movement across frames. This enlarged bbox $\hat{\mathbf{B}}_t$ is then applied to the surrounding frames $\{f_{t+k}\}_{k \neq 0}$ to ensure that the subject remains within the region of interest across all frames in the sequence:
\[
\hat{\mathbf{b}}_{t+k} = \hat{\mathbf{b}}_t \quad \text{for all } k \in \{-T, \dots, T\} \setminus \{0\}
\]
This strategy ensures capturing motion across frames effectively, enabling more accurate estimation of keypoints despite the subject moving in the video sequence.

\begin{figure*}[h]
  \centering
  \includegraphics[trim=9pt 0pt 230pt 50pt, clip, width=0.9 \linewidth]{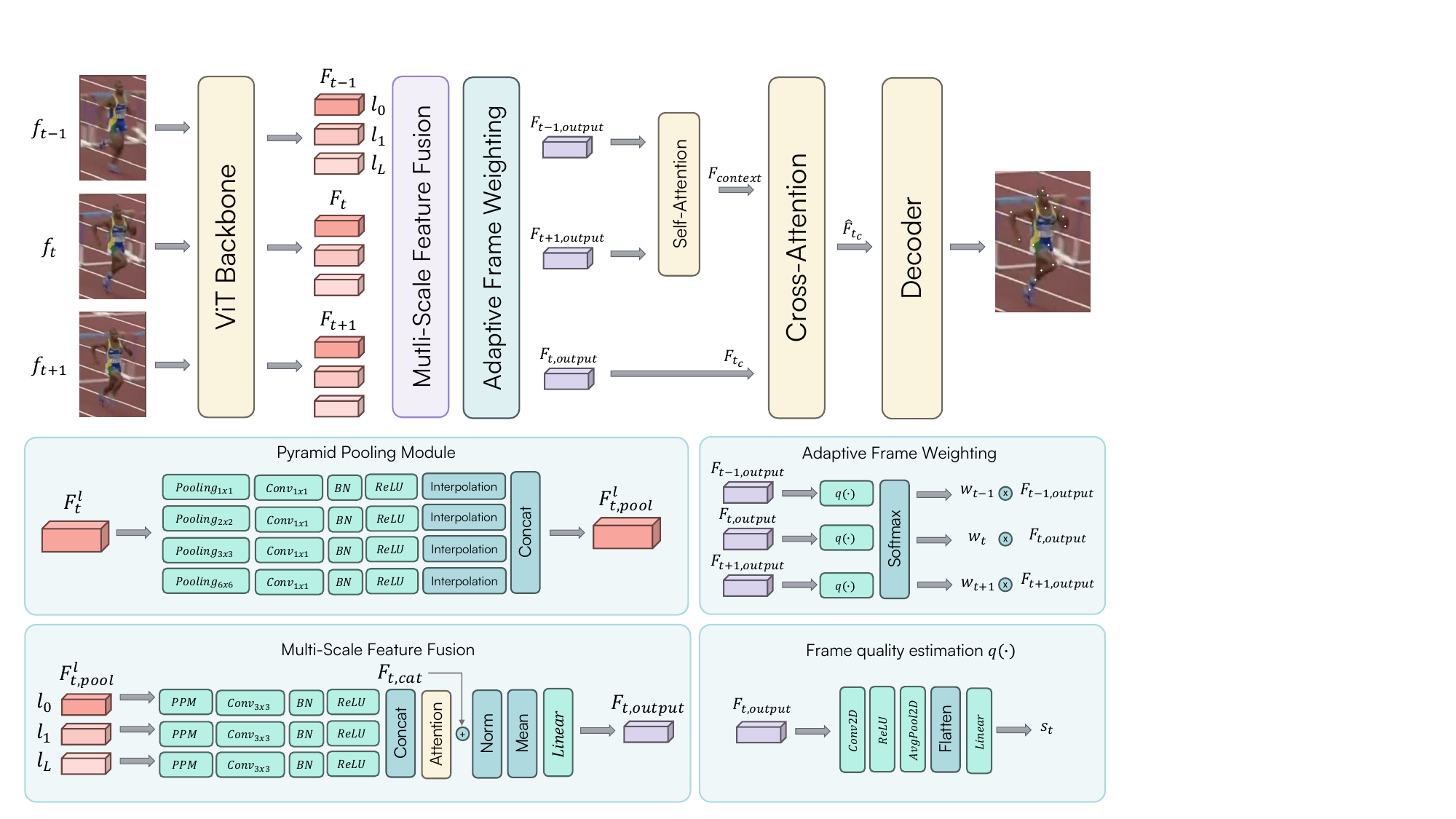}
    \caption{The architecture of the proposed Poseidon model for multi-frame human pose estimation. Poseidon builds on the ViTPose model by integrating an AFW mechanism, an MSFF module, and a Cross-Attention module. The AFW dynamically prioritizes frames based on their relevance within the sequence, while MSFF aggregates multi-scale features to enhance spatial understanding across different scales. Cross-attention facilitates information exchange between the central and context frames, capturing temporal dependencies critical for robust pose estimation in complex video sequences.}
  \label{fig:model_scheme}
\end{figure*}

The main pipeline of Poseidon consists of three stages: (1) Backbone and Decoder, (2) MSFF, (3) AFW, and (4) Cross-Attention Integration. Fig.~\ref{fig:model_scheme} shows the entire architecture and workflow of Poseidon. Subsequent sections go deeper into the explanation of the Poseidon components.

\subsection{Backbone and Decoder: ViTPose Adaptation}\label{sec:backbone_decoder}
In the Poseidon architecture, we adapt the backbone and decoder from the ViTPose model~\cite{xu2022vitpose}, known for its strong spatial feature extraction capabilities in single-frame pose estimation. Poseidon extends this model for multi-frame pose estimation by incorporating additional components while retaining the core ViTPose structure for adequate spatial understanding.

\subsubsection{Backbone (Encoder)}
The backbone, or encoder, is derived from the ViTPose architecture based on a Vision Transformer (ViT)\cite{dosovitskiy2021imageworth16x16words}. ViT operates by splitting input images into patches and applying a series of Transformer Blocks as shown in Fig.~\ref{fig:model_scheme}. This enables the model to capture local and global spatial information, crucial for precise keypoint localization.

\subsubsection{Decoder}
The decoder is adapted from the classic ViTPose decoder. The decoder is responsible for transforming the feature maps produced by the backbone into keypoint heatmaps, which represent the estimated positions of human joints in the input frames. It is composed of two deconvolution blocks, each of which contains one deconvolution layer followed by batch normalization\cite{ioffe2015batchnormalizationacceleratingdeep} and ReLU\cite{agarap2019deeplearningusingrectified}, then a convolution layer is utilized to get the
localization heatmaps for the keypoints, as shown in Fig.~\ref{fig:model_scheme}.

\subsection{Multi-Scale Feature Fusion (MSFF)}\label{sec:feature_fusion}

The MSFF module integrates features extracted from different levels of the backbone network. This integration is essential for accurately capturing fine-grained details (such as hand or foot keypoints) and high-level semantic information (such as whole body structure), often at different scales.
Human poses are inherently multi-scale, where smaller body parts (e.g., hands, feet) and larger structures (e.g., torso, limbs) must be detected simultaneously. Capturing complex poses, particularly in dynamic environments or occluded scenes, requires effectively integrating information across multiple scales. Multi-scale feature fusion addresses this by combining information from different feature levels, allowing the model to capture and reason about poses at varying granularities.

MSFF module is based on a Pyramid Pooling Module (PPM) inspired by ~\cite{He_2014}, which pools features at different scales and fuses them using a combination of convolutional layers and normalization. 

Let $\{F_t\}_{t=1}^T$ represent a sequence of T feature maps extracted from different layers $l$ (layer 9, layer 21 and final output), empirically chosen,  of the backbone network for each frame. Each feature map $F_t$ is processed through the MSFF mechanism to capture multi-level spatial information across scales.

\subsubsection{Pyramid Pooling Module (PPM)}

For each feature map $F_t$, the PPM applies adaptive pooling at multiple scales ($1\times1,2\times2,3\times3 \text{ and } 6\times6$) generating pooled features for each frame and layer. The pooled feature map at scale $p_i$ is denoted as:

\[
P^{(p_i)}(F^{(l)}_t) = \text{ReLU}(\text{BN}(\text{Conv}_{1\times1}(\text{AvgPool}{p_i}(F^{(l)}_t))))
\]
Here, the adaptive average pooling is followed by a $1\times1$ convolution, batch normalization, and ReLU activation. The convolution reduces the embedding dimension to a quarter of the original one, ensuring that the final dimensionality matches the initial feature map when concatenating the multi-scale feature maps. At the end, an interpolation layer brings back the H e W feature's dimension.

The multi-scale feature map for each layer and frame is obtained by concatenating the original feature map $F_t$ with the upsampled pooled feature maps:

\[
F_{t, \text{pool}}^{(l)} = \left[(P^{(p_1)}(F^{(l)}_t)), \dots, (P^{(p_m)}((F^{(l)}_t))) \right]
\]

This results in a multi-scale feature representation incorporating spatial context at varying resolutions while maintaining the same feature map dimensions as the original input.

\subsubsection{Feature Convolution and Normalization}

The concatenated multi-scale feature map $F_{t, \text{pool}}$ is processed through a convolutional layer to unify the feature dimensions. This step is followed by batch normalization and ReLU activation:

\[
F_{t, \text{conv}}^{(l)} = \text{ReLU}(\text{BN}(\text{Conv3x3}(F_{t, \text{pool}}^{(l)})))
\]

where $F_{t, \text{conv}}^{(l)} \in \mathbb{R}^{B \times C \times H \times W}$ represents the processed feature map at each scale for the t-th frame, in which $C$ is the output channel, $H$ and $W$ are the spatial dimension and $B$ the batch size.

\subsubsection{Attention-Based Fusion}

Once the multi-scale feature maps $
\left\{ F_{t, \text{conv}}^{(l)} \right\}_{t=1}^T, \quad \text{for } l = 0, 1, \dots, L
$ are obtained, we apply an attention mechanism to combine these features across layers and frames. 
The feature maps from different layers $l$ are concatenated along the channel dimension:

\[
F_{t, \text{cat}} = \left[F^{(0)}_{t, \text{conv}}, F^{(1)}_{t, \text{conv}}, \dots, F^{(L)}_{t, \text{conv}} \right]
\]

The concatenated feature map $F_{t, \text{cat}}$ is then passed through a multi-head attention module to capture dependencies between spatial locations across frames:

\[
\text{Attn}(F_{t, \text{cat}}) = \text{softmax}\left( \frac{F_{t, \text{cat}} W_Q \cdot (F_{t, \text{cat}} W_K)^T}{\sqrt{d_k}} \right) \cdot F_{t, \text{cat}} W_V
\]

where $W_Q, W_K, W_V$ are the learned query, key, and value matrices, and d\_k is the dimensionality of the key vectors.

To enhance the flow of information and stabilize the learning process, we introduce a residual connection between the original concatenated feature map $F_{t, \text{cat}}$ and the output of the attention mechanism:

\[
\hat{F}_{t, \text{cat}} = \text{Norm}(F_{t, \text{cat}} + \text{Attn}(F_{t, \text{cat}}))
\]

Here, \text{Norm} refers to layer normalization applied to the combined features. After the attention and normalization steps, we use a linear transformation to the fused features' original feature dimensionality.

Additionally, we apply an average pooling operation across the different layers after the attention fusion to ensure that information is effectively fused across all layers. This results in a unified multi-scale representation for each frame:

\[
F_{t, \text{fused}} = \text{Mean}\left( \hat{F}_{t, \text{cat}} \right) \quad \text{across layers}
\]

Finally, this fused representation is passed through a projection layer to ensure the output maintains the necessary dimensionality for downstream tasks.

\subsubsection{Final Projection}

After attention fusion, the final fused feature map is projected back to its original dimensions using a linear projection:

\[
F_{t, \text{final}} = \text{Linear}(F_{t, \text{fused}})
\]

The final output for each frame is reshaped back to its spatial dimensions, yielding the multi-scale fused feature map:

\[
F_{t, \text{output}} = \text{Reshape}(F_{t, \text{final}}) \in \mathbb{R}^{B \times C \times H \times W}
\]

This final output $F_{t, \text{output}}$ represents the multi-scale, multi-layer fused features for the t-th frame, incorporating both fine-grained details and high-level semantic information from the input sequence.

\subsection{Adaptive Frame Weighting}\label{sec:adaptive_frame_weighting}
The AFW is a component of the Poseidon architecture, designed to enhance the model's ability to focus on the most informative frames in a video sequence. This mechanism dynamically assigns weights to frames based on relevance, ensuring that the model pays more attention to frames containing critical pose information while down-weighting less informative or noisy frames.
In multi-frame pose estimation, not all frames are equally helpful for accurate keypoint localization. Some frames may contain occlusions, motion blur, or be less informative due to the lack of movement. Additionally, sequences with rapid movements or complex poses often have frames that offer more crucial information than others. The AFW mechanism addresses this by dynamically prioritizing frames contributing more to accurate pose estimation.

Let $\{F_{t, \text{output}}\}_{t=1}^T$ represent a sequence of T feature maps taken from the MSFF. The AFW mechanism assigns a weight $w_t$ to each feature map $F_{t, \text{output}}$, where the weight is computed using a learned quality estimation function $q(\cdot)$. The weighted feature map $WF$ for each frame is given by:

\[
{F}_{t, w} = w_t \cdot F_{t, \text{output}}
\]

The $q(F_{t, \text{output}})$ function is implemented as:

\[
q(\cdot) = \text{Linear(Flatten(AvgPool(ReLU(Conv3x3}(\cdot))))
\]

The weights $\{w_t\}_{t=1}^T$ are obtained normalizing$\{s_t\}_{t=1}^T$ coming from $q(\cdot)$ using a softmax function to ensure they sum to 1:

\[
w_t = \frac{e^{s_t}}{\sum_{k=1}^{T} e^{s_k}}
\]

where $s_t$ is the score assigned to frame $f_t$ by the frame quality estimator. This formulation ensures the model prioritizes frames with higher quality scores, focusing on more informative frames for accurate pose estimation.

The estimator analyzes each frame's feature map provided by the backbone to assign a quality score, which is then normalized to produce the final weights. The model applies these weights to the frames' feature maps before passing through subsequent layers.

This mechanism is particularly effective in handling occlusions, rapid movements, or other complexities in human pose estimation, allowing the model to adaptively focus on the most relevant parts of the video sequence.

\subsection{Cross-Attention}
The cross-attention module in Poseidon plays a key role in facilitating the exchange of information between the center frame and its surrounding context frames. Using a cross-attention mechanism, the model can better leverage the temporal context, improving its ability to capture relationships across frames, which is crucial for accurate pose estimation in video sequences where human motion is continuous and dynamic. This is particularly important in scenarios involving motion blur, occlusions, or fast movements where the pose information in a single frame might be ambiguous or incomplete.

The cross-attention module in Poseidon is implemented using multi-head attention from the Transformer architecture~\cite{dosovitskiy2021imageworth16x16words}. The feature map for the centre frame $F_{t_c}$ is treated as the query, and the feature maps of the context frames $F_{context} = \{F_{t_1}, \dots, F_{t_c-1}, F_{t_c+1}, \dots, F_T\}$ as keys and values. To improve the understanding of the context frames, before moving on to cross-attention, the context frames are applied to a self-attention layer.

We compute the attention weights for each spatial location in the centre frame by comparing them to all spatial locations in the context frames. The attention weights  $\alpha$  for each spatial location i in the center frame are computed as:

\[
\alpha_{i,j} = \frac{\exp\left( q(F_{t_c, i})^T k(F_{context, j}) / \sqrt{d_k} \right)}{\sum_{j=1}^{N} \exp\left( q(F_{t_c, i})^T k(F_{context, j}) / \sqrt{d_k} \right)}
\]

where:

\begin{itemize}
    \item $q(F_{t_c, i})$  is the query vector at location i in the center frame $F_{t_c}$
    \item $k(F_{context, j})$  is the key vector at location $j$ in the context frames $F_{context}$
    \item $d_k$  is the dimensionality of the key vectors,
    \item $N$ is the number of spatial locations across all context frames.
\end{itemize}

The attention weights are used to compute a weighted sum of the values from the context frames  $v(F_{context, j})$, which updates the feature map for the center frame:

\[
\hat{F}_{t_c, i} = \sum{j=1}^{N} \alpha_{i,j} v(F_{context, j})
\]
Here,  $\hat{F}_{t_c}$  is the updated feature map for the center frame after incorporating information from the context frames and is then passed to the decoder to generate the final keypoint predictions for the human pose estimation.

\section{Experiments}
In our experiments, we evaluated the proposed Poseidon model using the large-scale PoseTrack21~\cite{posetrack21}, PoseTrack18~\cite{posetrack18} and Sub-JHMDB~\cite{subjhmdb} for multi-frame human pose estimation. 

\subsection{Experimental Settings}

\paragraph{Dataset}
The PoseTrack18 dataset~\cite{posetrack18} is a large-scale benchmark for video-based human pose estimation and articulated multi-person tracking. It contains 550 video sequences with 66,374 frames. The dataset includes over 153,000 pose annotations and 23,000 labelled frames, capturing complex scenes with multiple people engaged in dynamic activities. PoseTrack18 addresses challenges like body motion, occlusions, person identity tracking, and varying person scales across videos. 

The PoseTrack21 dataset~\cite{posetrack21} is designed for multi-frame pose estimation, person search, multi-object tracking, and multi-person pose tracking in real-world video sequences. It includes over 420,000 bounding box annotations and 177,000 human pose annotations, making it one of the most extensive datasets for these tasks. It contains 593 videos for training and 170 videos for validation.
We evaluate the performance in PoseTrack18~\cite{posetrack18} and Posetrack21~\cite{posetrack21} dataset using the Mean Average Precision (mAP), which is the mean over the Average Precision (AP) of each joint. 

The Sub-JHMDB dataset~\cite{subjhmdb} is a subset of the JHMDB dataset, commonly used for evaluating human pose estimation in the context of activity recognition. It includes 316 video clips, each containing complex human actions, and provides annotations for 15 human joints across various activities. The dataset is challenging due to its wide variety of dynamic poses, occlusions, and background diversity. For this dataset, we evaluate performance using the Percentage of Correct Keypoints (PCK) metric, calculated with a threshold of 0.2, which considers a keypoint correctly predicted if it falls within 20\% of the person's torso length from the ground truth.

\paragraph{Backbone}
We experimented with different backbones based on the ViTPose architecture: ViT-S, ViT-B, ViT-L, and ViT-H. These backbones were pre-trained on human pose estimation tasks to provide strong spatial representations. The model processes images of size $288 \times 384$.
\paragraph{Training Configuration}
After some preliminary trials, the model was trained with a batch size of 16. We used eight joints with a probability of 0.3 for half-body augmentation to apply the augmentation. The rotation factor was set to $\pm 45^\circ$, and the scale factor was set to [0.35, 0.35]. The model was trained for 20 epochs using the AdamW optimizer. The initial learning rate was set to $5 \times 10^{-6}$, with a weight decay of 0.1. We used the StepLR learning rate scheduler with a step size of 5 and a learning rate decay factor of 0.5. The temporal window for the multi-frame input consisted of 2 previous frames and 2 next frames. the bounding box around each person was enlarged by a factor of $\delta = 1.25$ to capture a wider context around the subject. The model training process also includes using Automatic Gradient Scaling and Automatic Mixed Precision, which helps optimize GPU memory usage and speeds up training while maintaining numerical stability. 
The model was trained on an NVIDIA A100 GPU with 80GB of memory.

\paragraph{Loss Function}
The loss function used in our model is the Mean Squared Error loss, aiming to minimize the distance between the predicted and ground truth heatmaps for each joint. The ground truth heatmaps are generated using 2D Gaussian distributions centered at the joint locations. 

\subsection{Comparison with State-of-the-art Approaches}

In Tab \ref{tab:comparison_posetrack18} and \ref{tab:comparison_posetrack21}, we compare our Poseidon model and other state-of-the-art approaches on the PoseTrack21~\cite{posetrack21}and Posetrack18~\cite{posetrack18} dataset. The tables include baseline methods as provided in the original dataset paper \cite{Rafi_2020_corrtrack, Bergmann_2019_tracktor, posetrack21}. Additionally, we report high-performing methods such as DetTrack~\cite{dettrack}, PoseWarper~\cite{posewarped}, DCPose~\cite{liu2021deep}, FAMI-Pose~\cite{liu2022temporalfeaturealignmentmutual_famipose}, DiffPose~\cite{feng2023diffpose}, DSTA~\cite{he2024video}, TDMI-ST~\cite{feng2023mutual}, and JM-Pose~\cite{wu2024joint}.

\paragraph{Results on Posetrack18} As shown in Table \ref{tab:comparison_posetrack18}, our proposed Poseidon model demonstrates superior performance compared to other state-of-the-art methods on the PoseTrack18 dataset~\cite{posetrack18}. Notably, Poseidon achieves significant improvements in mAP, reaching 87.8. This represents a substantial enhancement over the previous top performer, JM-Pose~\cite{wu2024joint}, which scored 84.1. Our model shows particular strength in challenging keypoint estimations such as the wrist and ankle, where Poseidon achieves 86.3 and 87.2, respectively, surpassing other methods by considerable margins. 

\begin{table}
\resizebox{0.48\textwidth}{!}{%
  \centering
  \renewcommand{\arraystretch}{1.4} 
  \begin{tabular}{@{}l|ccccccc|c@{}}
    \toprule
    Method & Head & Shoulder & Elbow & Wrist & Hip & Knee & Ankle & \textbf{Mean} \\
    \midrule
    PoseWarper~\cite{posewarped} & 79.9 & 86.3 & 82.4 & 77.5 & 79.8 & 78.8 & 73.2 & 79.7\\
    DCPose~\cite{liu2021deep} & 84.0 & 86.6 & 82.7 & 78.0 & 80.4 & 79.3 & 73.8 & 80.9 \\
    DetTrack~\cite{dettrack} & 84.9 & 87.4 & 84.8 & 79.2 & 77.6 & 79.7 & 75.3 & 81.5 \\
    FAMI-Pose~\cite{liu2022temporalfeaturealignmentmutual_famipose} & 85.5 & 87.8 & 84.2 & 79.2 & 81.4 & 81.1 & 74.9 & 82.2 \\
    
    DiffPose~\cite{feng2023diffpose} & 85.0 & 87.7 & 84.3 & 81.5 & 81.4 & 82.9 & 77.6 & 83.0 \\
    DSTA~\cite{he2024video} & 85.9 & 88.8 & 85.0 & 81.1 & 81.5 & 83.0 & 77.4 & 83.4 \\
    TDMI-ST~\cite{feng2023mutual} & 86.7 & 88.9 & 85.4 & 80.6 & 82.4 & 82.1 & 77.6 & 83.6 \\
    JM-Pose~\cite{wu2024joint} & 86.6 & 88.7 & 86.0 & 81.6 & 83.3 & 83.2 & 78.2 & 84.1 \\
    \hline
    \textbf{Poseidon (Ours)} & \textbf{88.8} & \textbf{91.4} & \textbf{88.6} & \textbf{86.3} & \textbf{83.3} & \textbf{88.8} & \textbf{87.2} & \textbf{87.8} \\
    
    \bottomrule
  \end{tabular}}
  \caption{Quantitative results on the PoseTrack18 dataset.}
  \label{tab:comparison_posetrack18}
\end{table}

\paragraph{Results on Posetrack21} 
Our Poseidon model outperforms all other methods, achieving a significant improvement in terms of overall accuracy. Notably, the mAP of our method is 88.3, representing an increase over the previous best-performing method, JM-Pose~\cite{wu2024joint}, which achieved 84.0.

Poseidon also shows marked improvements in the wrist and ankle keypoints when focusing on specific joints. Poseidon reaches an mAP of 85.8 for the wrist joint, compared to JM-Pose~\cite{wu2024joint}, which had 82.5. Similarly, Poseidon achieves an mAP of 85.7 for the ankle joint over JM-Pose's 78.5.

\begin{table}
\resizebox{0.48\textwidth}{!}{%
  \centering
  \renewcommand{\arraystretch}{1.4} 
  \begin{tabular}{@{}l|ccccccc|c@{}}
    \toprule
    Method & Head & Shoulder & Elbow & Wrist & Hip & Knee & Ankle & \textbf{Mean} \\
    \midrule
    Tracktor++ w. poses~\cite{Bergmann_2019_tracktor, posetrack21} & - & - & - & - & - & - & - & 71.4\\
    CorrTrack~\cite{Rafi_2020_corrtrack, posetrack21} & - & - & - & - & - & - & - & 72.3 \\
    CorrTrack w. ReID~\cite{Rafi_2020_corrtrack, posetrack21} & - & - & - & - & - & - & - & 72.7 \\
    Tracktor++ w. corr.~\cite{Bergmann_2019_tracktor, posetrack21}  & - & - & - & - & - & - & - & 73.6 \\
    DCPose~\cite{liu2021deep} & 83.2 & 84.7 & 82.3 & 78.1 & 80.3 & 79.2 & 73.5 & 80.5 \\
    FAMI-Pose~\cite{liu2022temporalfeaturealignmentmutual_famipose} & 83.3 & 85.4 & 82.9 & 78.6 & 81.3 & 80.5 & 75.3 & 81.2 \\
    DiffPose~\cite{feng2023diffpose} & 84.7 & 85.6 & 83.6 & 80.8 & 81.4 & 83.5 & 80.0 & 82.9 \\
    DSTA~\cite{he2024video} & 87.5 & 87.0 & 84.2 & 81.4 & 82.3 & 82.5 & 77.7 & 83.5 \\
    TDMI-ST~\cite{feng2023mutual} & 86.8 & 87.4 & 85.1 & 81.4 & 83.8 & 82.7 & 78.0 & 83.8 \\
    JM-Pose~\cite{wu2024joint} & 85.8 & 88.1 & 85.7 & 82.5 & 84.1 & 83.1 & 78.5 & 84.0 \\
    \hline
    \textbf{Poseidon (Ours)} & \textbf{92.2} & \textbf{90.8} & \textbf{88.3} & \textbf{85.8} & \textbf{85.5} & \textbf{87.7} & \textbf{85.7} & \textbf{88.3} \\
    
    \bottomrule
  \end{tabular}}
  \caption{Quantitative results on the PoseTrack21 dataset.}
  \label{tab:comparison_posetrack21}
\end{table}

\paragraph{Results on Sub-JHMDB}
To evaluate the generalization capability of our Poseidon model, we tested the model trained on the PoseTrack21 dataset directly on the Sub-JHMDB dataset~\cite{subjhmdb} without any additional fine-tuning or retraining. This dataset consists of three distinct splits, and we report the mean Percentage of Correct Keypoints (PCK) over these splits with a threshold of 0.2.

\begin{table}
\resizebox{0.48\textwidth}{!}{%
  \centering
  \renewcommand{\arraystretch}{1.4} 
  \begin{tabular}{@{}l|ccccccc|c@{}}
    \toprule
    Method & Head & Shoulder & Elbow & Wrist & Hip & Knee & Ankle & \textbf{Mean} \\
    \midrule
    Thin-slicing Net~\cite{thin-slicing} & 97.1 & 95.7 & 87.5 & 81.6 & 98.0 & 92.7 & 89.8 & 92.1\\
    LSTM PM~\cite{lstmpm} & 98.2 & 96.5 & 89.6 & 86.0 & 98.7 & 95.6 & 90.0 & 93.6\\
    DKD~\cite{dkd} & 98.3 & 96.6 & 90.4 & 87.1 & 99.1 & 96.0 & 92.9 & 94.0\\
    SimplePose~\cite{simplepose} & 97.5 & 97.8 & 91.1 & 86.0 & 99.6 & 96.8 & 92.6 & 94.4\\
    K-FPN~\cite{k-fpn} & 95.1 & 96.4 & 95.3 & 91.3 & 96.3 & 95.6 & 92.6 & 94.7\\
    Motion Adaptive~\cite{motionadaptive} & 98.2 & 97.4 & 91.7 & 85.2 & 99.2 & 96.7 & 92.2 & 94.7\\
    DeciWatch~\cite{zeng2022deciwatch} & \textbf{99.9} & \textbf{99.8} & \textbf{99.6} & \textbf{99.7} & \textbf{98.6} & \textbf{99.6} & \textbf{96.6} & \textbf{98.9}\\
    \hline
    \textbf{Poseidon (Ours)} & 99.5 & 99.0 & 95.9 & 93.3 & 98.5 & 98.3 & 96.5 & 97.3 \\
    \bottomrule
  \end{tabular}}
  \caption{Quantitative results on the Sub-JHMDB dataset.}
  \label{tab:comparison_subjhmdb}
\end{table}

The results in Table~\ref{tab:comparison_subjhmdb} demonstrate the strong generalization capability of our Poseidon model on the Sub-JHMDB dataset, reaching a mean PCK of 97.3. This result places it very competitively among existing methods, outperforming several state-of-the-art models like Thin-slicing Net~\cite{thin-slicing}, LSTM PM~\cite{lstmpm}, DKD~\cite{dkd}, Motion Adaptive~\cite{motionadaptive}, K-FPN~\cite{k-fpn}, and SimplePose~\cite{simplepose}. Compared to DeciWatch~\cite{zeng2022deciwatch}, which achieves a mean PCK of 98.9, Poseidon falls short by a small margin but still maintains competitive performance without additional fine-tuning on the Sub-JHMDB dataset.

\subsection{Ablation Study}
Different backbones are employed as shown in Tab~\ref{tab:comparison_posetrack21_different_backbone}, demonstrating performance improvements when using different backbones on the PoseTrack21 dataset. As the model's capacity increases from ViT-S to ViT-H~\cite{dosovitskiy2021imageworth16x16words}, there is a clear trend of higher keypoint accuracy across all body parts, reflecting the positive impact of more extensive and expressive backbones. The ViT-H backbone achieves the highest mean performance of 88.3, significantly outperforming ViT-S by 6.6 points. The columns \#BP and \#TP represent the number of parameters (in millions) within the backbone and the total parameters of each model configuration, respectively. In particular, the ViT-L backbone reaches 87.3 of mAP, which is only 1 point lower than ViT-H, yet achieves this performance with approximately half the total parameters, as shown in Tab~\ref{tab:comparison_posetrack21_different_backbone}. This highlights ViT-L as an effective option that balanced option between performance and model size.
Notably, DiffPose~\cite{feng2023diffpose} and DSTA~\cite{he2024video} also utilize the ViT-H backbone but achieve mean accuracies of 82.9 and 83.5, respectively, which are substantially lower than the 84.6 achieved by Poseidon with ViT-B as the backbone. This indicates that Poseidon's architecture is more efficient and effective, leveraging the backbone capabilities better than these other approaches.
Other approaches in Tab.~\ref{tab:comparison_posetrack21} employ HRNet-w48~\cite{sun2019deep_hrnet}, which has 77.5M parameters. While HRNet-w48~\cite{sun2019deep_hrnet} is comparable in size to ViT-B, its performance lags, with TDMI-ST~\cite{feng2023mutual} and JM-Pose~\cite{wu2024joint} reaching mean accuracies of 83.8 and 84.0, respectively, still falling short of Poseidon's performance. This suggests that Poseidon's use of ViT-B provides a more advantageous balance of model capacity and parameter efficiency than HRNet-based methods despite the HRNet's established effectiveness in pose estimation tasks.

\begin{table}
\resizebox{0.48\textwidth}{!}{%
  \centering
  \renewcommand{\arraystretch}{1.4} 
  \begin{tabular}{@{}ccc|ccccccc|c@{}}
    \toprule
    Backbone& \#BP (M) & \#TP (M) & Head & Shoulder & Elbow & Wrist & Hip & Knee & Ankle & \textbf{Mean} \\
    \midrule
    ViT-S  & 21.7 & 29.6 & 88.3 & 86.5 & 82.3 & 77.0 & 79.4 & 79.0 & 76.3 & 81.7 \\
    ViT-B  & 85.8 & 110.0 & 89.7 & 88.3 & 84.9 & 80.5 & 82.6 & 83.1 & 80.6 & 84.6 \\
    ViT-L  & 343.6 & 303.3 & 91.1 & 90.1 & 87.6 & 84.4 & 85.4 & 86.0 & 84.7 & 87.3 \\
    ViT-H  & 630.9 & 691.5 & 92.2 & 90.8 & 88.3 & 85.8 & 85.5 & 87.7 & 85.7 & 88.3 \\
    \bottomrule
  \end{tabular}}
  \caption{Quantitative results on the PoseTrack21 validation set with different backbone.}
  \label{tab:comparison_posetrack21_different_backbone}
\end{table}

\begin{table}
\resizebox{0.48\textwidth}{!}{%
  \centering
  \renewcommand{\arraystretch}{1.4} 
  \begin{tabular}{@{}c|ccc|ccccccc|cc@{}}
    \toprule
    Methods & Cross-Att\textsuperscript{1} & AFW & MSFF & Head & Shoulder & Elbow & Wrist & Hip & Knee & Ankle & \textbf{Mean}\\
    \midrule
    (a)         & & & & 91.1 & 90.4 & 87.4 & 84.5 & 85.2 & 85.7 & 83.1 & 87.1\\
    (b)         & \checkmark & \checkmark & & 90.9 & 90.7 & 88.0 & 84.6 & 85.5 & 86.5 & 84.5 & 87.5 \textcolor{ForestGreen}{(+0.4)} \\
    (c)         & \checkmark & & \checkmark & 91.7 & 91.1 & 88.2 & 85.4 & 85.5 & 87.1 & 85.3 & 88.0 \textcolor{ForestGreen}{(+0.9)}\\
    Poseidon    & \checkmark & \checkmark & \checkmark & 92.2 & 90.8 & 88.3 & 85.8 & 85.5 & 87.7 & 85.7 & 88.3 \textcolor{ForestGreen}{(+1.2)} \\
    \bottomrule
  \end{tabular}
  }
  \caption{Ablation study on the PoseTrack21 validation set}
  \label{tab:ablation_study_posetrack21}
\end{table}

The ablation study in Table \ref{tab:ablation_study_posetrack21} highlights the impact of various components on the Poseidon model's performance. Experiment (a), which uses only the ViTPose architecture~\cite{xu2022vitpose} with the central frame, serves as the baseline with a mean accuracy of 87.1, showcasing the limitations of single-frame reliance, especially for joints requiring temporal context. Adding cross-attention and AFW in experiment (b) increases accuracy to 87.5 (+0.4 points), enhancing frame prioritization and information exchange for better consistency, notably improving elbows and knees.
Experiment (c) incorporates cross-attention with MSFF, raising accuracy to 88.0 (+0.9 points) by effectively integrating multi-scale features for precise joint localization. The complete Poseidon architecture, combining AFW, MSFF, and cross-attention, achieves the highest mean accuracy of 88.3 (+1.2 points), significantly enhancing performance for complex poses, especially at challenging joints like wrists and ankles.

\section{Conclusion and Future Work}
In this paper, we introduced Poseidon, a novel architecture for human pose estimation exploiting a multi-frame approach building upon the strengths of the ViTPose model and adapting it for temporal video analysis. By integrating an AFW mechanism, an MSFF module, and a Cross-Attention module, Poseidon effectively captures spatial and temporal information to improve keypoint localization in video sequences.

Our comprehensive evaluation of the PoseTrack18 and PoseTrack21 data\-sets demonstrates that Poseidon achieves state-of-the-art performance, outperforming existing methods by significant margins. The generalization capability of Poseidon was further validated by testing it on the Sub-JHMDB dataset without any fine-tuning.

The ablation studies underscore the importance of these architectural components, showing that each contributes to enhancing the overall pose estimation performance. Poseidon achieves competitive accuracy and offers a balanced option between performance and model size with the different backbone sizes. It is well-suited for real-world applications where precision and robustness are paramount.

Future work will explore extending Poseidon to more generalized tasks, such as multi-person tracking and action recognition, and optimizing the model for deployment on edge devices to support real-time applications. By continuing to build on the concepts introduced in Poseidon, we aim to advance the field of human pose estimation, enabling more accurate and context-aware analysis of human movement in diverse environments

\clearpage
\bibliographystyle{elsarticle-num} 
\bibliography{bib.bib}
\end{document}